\documentclass{article}

\usepackage[preprint]{neurips_2026}

\usepackage[utf8]{inputenc}
\usepackage[T1]{fontenc}
\usepackage{hyperref}
\usepackage{url}
\usepackage{booktabs}
\usepackage{amsfonts}
\usepackage{nicefrac}
\usepackage{microtype}
\usepackage{xcolor}
\usepackage{amsmath}
\usepackage{amssymb}
\usepackage{graphicx}
\usepackage{algorithm}
\usepackage{algpseudocode}
\usepackage{enumitem}

\title{Align-RAG: Alignment Is All You Need for TSFM In-Context Learning}

\author{%
  \textbf{Mohammad Asadi}$^{1}$\quad
  \textbf{Soheil Hor}$^{2}$\quad
  \textbf{Bardiya Akhbari}$^{2}$\\[2pt]
  \textbf{Jack W.~O'Sullivan}$^{1}$\quad
  \textbf{Tahoura Nedaee}$^{1}$\quad
  \textbf{Layne C.~Price}$^{2}$\\[2pt]
  \textbf{Raviteja Anantha}$^{2}$\quad
  \textbf{Euan Ashley}$^{1}$\quad
  \textbf{Ehsan Adeli}$^{1}$\\[8pt]
  $^{1}$Stanford University\qquad$^{2}$Amazon\\[4pt]
  \texttt{masadi@stanford.edu}%
}

\begin{document}

\maketitle

\begin{abstract}
Retrieval-augmented forecasting promises to adapt frozen Time Series Foundation Models (TSFMs) to new domains without fine-tuning, but recent methods typically rely on learned fusion modules, i.e., trained adapters that merge retrieved examples into the backbone's forecast, based on the assumption that frozen backbones cannot dynamically incorporate retrieved context on their own. We show this assumption is unnecessary. We introduce \textbf{Align-RAG}, a training-free method that applies a closed-form per-pair amplitude rescaling and integer-lag phase shift to retrieved past--future windows before they enter a frozen backbone's context. With no learned parameters, Align-RAG outperforms the state-of-the-art trained retrieval adapter on a frozen Chronos-Bolt on all seven datasets of the standard benchmark (avg $-3.75\%$ MSE), showing that the gains previously attributed to learned fusion are recoverable without any training. Align-RAG further improves zero-shot MSE on four additional frozen TSFMs with various architectures by $2.5\%$ to $13.7\%$ per backbone with no per-backbone tuning. To probe why alignment helps, we compare the frozen backbone's prediction shift under aligned demonstrations to the closed-form ridge prediction shift on the same pairs. We find that aligned demonstrations induce prediction shifts that track a closed-form ridge predictor on the same pairs, with a future-shuffle control ruling out a futures-averaging account. Together, these results indicate that frozen TSFMs already support dynamic in-context use of retrievals, and that closed-form alignment should be the default baseline for retrieval-augmented forecasting before any fusion module is trained. Code available at: \url{https://github.com/masadi-99/align-rag}

\end{abstract}

\section{Introduction}
\label{sec:intro}

\begin{figure}[t]
\centering
\includegraphics[width=\linewidth]{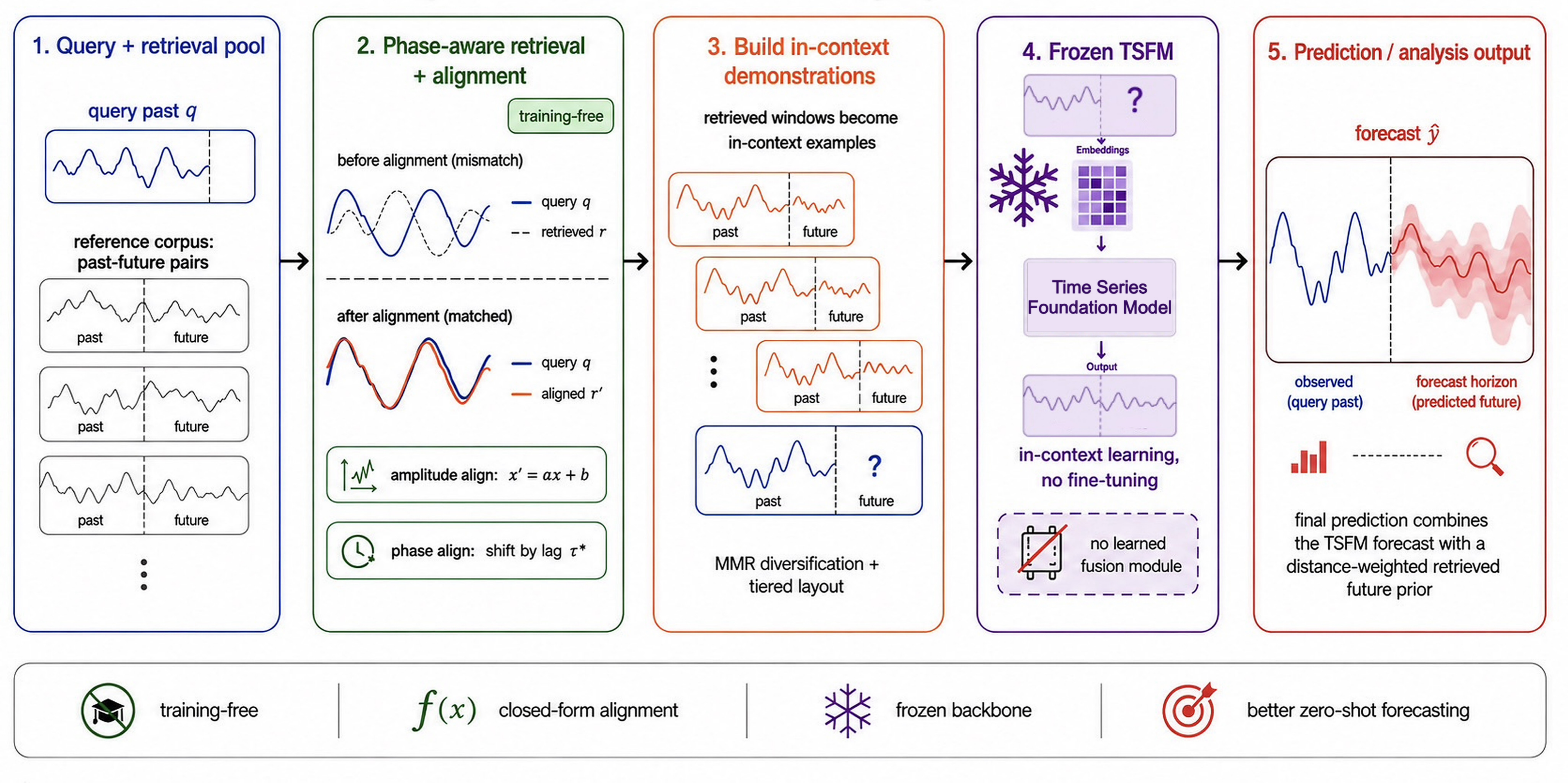}
\caption{Align-RAG overview. (1) For each query past $\mathbf{q}$, retrieve
top-$K$ past--future windows from a reference corpus. (2) Apply two
closed-form, training-free transforms per pair so that each
retrieved past matches the query while the past--future correspondence is
preserved. (3) Pack the aligned pairs as in-context demonstrations using MMR
diversification and a tiered token-budget layout. (4) A single forward pass
through a frozen Time Series Foundation Model. No learned fusion module is
introduced. (5) The final forecast combines the model's in-context output
with a distance-weighted prior built from the retrieved aligned futures.}
\label{fig:overview}
\end{figure}

Time series forecasting underpins decision-making in domains as
varied as energy management, finance, healthcare, and climate
science. Time Series Foundation Models (TSFMs) such as
Chronos~\citep{ansari2024chronos},
Chronos-2~\citep{shchur2025chronos2},
TimesFM~\citep{das2024timesfm}, Moirai~\citep{woo2024moirai},
and Toto~\citep{cohen2024toto}
have established a zero-shot forecasting paradigm by learning
shared temporal representations across heterogeneous corpora.
Once pretrained, a frozen TSFM can be applied to unseen series
without further training, but its accuracy on a target distribution
depends on whatever the pretraining corpus exposed.

Retrieval-augmented generation~\citep{lewis2020rag} addresses an
analogous gap for large language models by providing relevant
context at inference time. The corresponding move for time series
is to retrieve past--future windows from a reference set whose
dynamics resemble the query and supply them to the model as
additional context. For example, a hospital deploying a frozen TSFM might
retrieve patterns from related ICU traces and an energy operator might
retrieve consumption windows from similar load regimes. Done well,
this offers dynamic adaptation without labelled data or per-domain
fine-tuning.

Recent retrieval-augmented forecasting on frozen TSFMs has
converged on a single design pattern: pair the backbone with a
learned fusion module that mediates between retrievals and the
model's representation. TS-RAG (Neurips 2025)~\citep{ning2025tsrag}, the state-of-the-art in retrieval-augmented forecasting for time series foundation models to the best of our knowledge, states the
position cleanly, claiming that pretrained TSFMs ``lack inherent
mechanisms for domain adaptation, as they cannot incorporate
external contextual knowledge dynamically'', and trains an Adaptive
Retrieval Mixer to fuse encoded retrievals with the backbone.
TimeRAF~\citep{xie2024timeraf} trains an end-to-end retriever and
a Channel Prompting module; RATD~\citep{liu2024ratd} trains a
reference-modulated attention block inside a diffusion denoiser;
RAFT~\citep{yang2025raft} trains projection and prediction heads on
top of the retrieved patterns;
TimesFM-ICF~\citep{das2024incontextft} re-pretrains the foundation
model itself to consume in-context retrieval examples. The
underlying assumption is uniform: a learned ad-hoc element is
required to make a frozen TSFM benefit from retrieved context. We
challenge this assumption.

We introduce Align-RAG, a training-free retrieval-augmented
forecasting method that extends a frozen TSFM by aligning retrieved
past--future windows to the query in amplitude and phase before
they enter the backbone's context. Given a query window, Align-RAG
retrieves nearest neighbours from a reference set of (past, future)
pairs and applies a closed-form per-pair amplitude rescaling and
phase shift so that each retrieved past matches the query. The
aligned pairs are then provided as in-context demonstrations to
the backbone, whose parameters remain unchanged.

Align-RAG outperforms TS-RAG on a frozen Chronos-Bolt backbone on
$7/7$ datasets in MSE and $6/7$ in MAE across the TS-RAG benchmark
protocol (ETTh1, ETTh2, ETTm1, ETTm2, Weather, Exchange,
Electricity; input length $512$, horizon $64$), with an average
MSE reduction of $3.75\%$ and gains significant under a paired
moving-block bootstrap on $5$ of $6$ small-dataset comparisons.
The same procedure transfers across four additional frozen TSFMs
(Chronos-2, TimesFM-2.0, Moirai, Toto), improving zero-shot MSE by
between $2.5\%$ and $13.7\%$ per backbone, with the largest gains
on Moirai (avg $-13.7\%$, single cells reaching $-30\%$ on ETTm1
and $-20\%$ on Weather), and without any per-backbone tuning. The comparison is
non-trivial: TS-RAG's Adaptive Retrieval Mixer is trained on $26$
million context--horizon pairs sampled from Chronos's
pretraining corpus~\citep{ning2025tsrag}, while Align-RAG
introduces no learned parameters at all.

A $2{\times}2$ ablation isolates where the gain comes from. Holding
the backbone and benchmark fixed, we cross retriever (trained
TS-RAG retriever vs.\ random retrieval) with alignment (on vs.\
off). Switching from the trained retriever to a random retriever,
with alignment on, costs $0.84$ percentage points of MSE on average
and still beats the trained TS-RAG mixer on $4$ of $7$ datasets.
Turning alignment off, with either retriever, costs more than $20$
percentage points of MSE. The ranking of retrieved windows accounts
for under one point of the headline; the alignment of the windows
that are retrieved accounts for the rest.

We also probe the mechanism behind this gain. For each test window
we compute the cosine similarity $\rho_{\rm GD}$ between the frozen
backbone's prediction shift, induced by replacing zero-shot context
with aligned demonstrations, and the closed-form ridge prediction
shift on the same demonstration pairs. On Chronos-Bolt,
$\rho_{\rm GD}$ rises from $0.30$ without alignment to $0.45$ with
alignment, and a paired moving-block bootstrap excludes zero on
$7/7$ datasets. As a causal control, we shuffle the futures of the
retrieved pairs while leaving the pasts and the alignment intact.
On ETTm1, the $4.2\%$ MSE reduction from alignment flips to a
$12.2\%$ MSE increase under the future shuffle, in the direction
the closed-form ridge predicts on the same shuffled pairs, ruling
out an explanation in which the backbone is averaging over
retrieved futures and ignoring the past--future relationship.

\paragraph{Contributions.}
\begin{itemize}[noitemsep,topsep=2pt]
\item Align-RAG, a training-free retrieval-augmented forecasting
method that aligns retrieved past--future windows to the query in
amplitude and phase before they enter a frozen TSFM's context,
without modifying the backbone or training any new component.
\item Evaluation against the trained TS-RAG mixer on the standard
seven-dataset benchmark with a frozen Chronos-Bolt
(Section~\ref{sec:main_result}) and across four further frozen
TSFMs (Section~\ref{sec:cross_backbone}).
\item A behavioural mechanism analysis
(Section~\ref{sec:why_phaserag}): under our closed-form alignment,
the frozen backbone's prediction shift tracks the closed-form
ridge prediction shift on the same demonstrations, and a
future-shuffle causal control confirms the model is regressing on
the (past, future) pairs rather than averaging over their futures.
\end{itemize}

\section{Related work}
\label{sec:related}

\paragraph{Retrieval-augmented forecasting for TSFMs.}
A growing body of work attaches trained components to a frozen
TSFM to consume retrievals.
TS-RAG~\citep{ning2025tsrag} trains an Adaptive Retrieval Mixer
on top of frozen Chronos-Bolt~\citep{ansari2024chronos};
TimeRAF~\citep{xie2024timeraf} trains an end-to-end retriever and
a Channel Prompting module; RAFT~\citep{yang2025raft} trains
projection and prediction heads; RATD~\citep{liu2024ratd} trains a
diffusion denoiser with a learned reference-modulated attention block;
TimeRAG~\citep{yang2024timerag} inserts a learned reprogramming
layer; Advanced RAF~\citep{tire2024raf} fine-tunes the backbone
end-to-end (while Naive RAF, although a training-free variant, is outperformed substantially by TS-RAG\citep{ning2025tsrag}). Each of
these methods relies on a learned ad-hoc element interposed between
the retrieval and the backbone. We compare numerically against
TS-RAG, which is the most directly comparable and the strongest
trained-fusion baseline at the time of writing; we adopt its
protocol throughout for fair comparison. The corresponding move
on the LLM side has typically been simpler. Methods such as
REPLUG~\citep{shi2024replug},
$k$NN-LM~\citep{khandelwal2020knn}, and In-Context
RALM~\citep{ram2023incontext} concatenate or interpolate retrieved
context with a frozen LM directly.

\paragraph{In-context learning as implicit regression.}
A line of theoretical work frames in-context learning as implicit
regression. \citet{garg2022whatcan} show that transformers trained
on linear functions match the least-squares estimator;
\citet{akyurek2023learning} and \citet{vonoswald2023} demonstrate
that the algorithm such transformers implement is consistent with
gradient descent or closed-form ridge regression on the in-context
demonstrations. \citet{das2024incontextft} extends a related lens
to TSFMs by re-pretraining the foundation model to consume
in-context retrieval examples. We use this body of work as a
behavioural reference: in Section~\ref{sec:why_phaserag} we compare
the frozen TSFM's prediction shift under our alignment to the
closed-form ridge predictor's shift on the same demonstrations.

\paragraph{Phase, amplitude, and structural similarity in time series.}
Dynamic time warping~\citep{cuturi2017softdtw} and its shape
variants~\citep{zhao2018shapedtw} compare misaligned series;
$k$-Shape~\citep{paparrizos2015kshape} clusters by phase-invariant
Pearson similarity; the matrix profile~\citep{yeh2016matrixprofile}
indexes motifs under $z$-normalised distance. These methods treat
amplitude and phase as nuisance directions and remove them before
downstream learning.

\section{Method: Align-RAG}
\label{sec:method}

Align-RAG operates on each retrieved past--future window, before
tokenization, with two closed-form transforms. The first rescales
the retrieved past so its first and second moments agree with the
query's, through a regularised affine fit. The second shifts the
retrieved past in time so its principal lag with respect to the
query is zero, as an integer-sample shift chosen by
cross-correlation. Both transforms have the same future-side
counterparts: the same affine map and the same lag, applied to the
retrieved future so the $(\mathbf{p}_i,\mathbf{f}_i)$ pair stays
consistent. Neighbour diversification, packing into the
backbone's context budget, future blending, and a no-demonstration
second forward pass are further training-free components that
complement the alignment and improve zero-shot accuracy.

Let $\mathbf{q}\in\mathbb{R}^{S}$ denote the query past of length
$S$ ($S{=}512$ for Chronos-Bolt) and $\mathbf{y}\in\mathbb{R}^{H}$
its target future of horizon $H$. A retrieval index returns a pool
$\mathcal{P}=\{(\mathbf{p}_i,\mathbf{f}_i,d_i)\}_{i=1}^{20}$ of
(past, future, distance) triples per query channel. We write
$f_\theta(C)$ for the frozen TSFM's forecast given a context $C$.

\subsection{Amplitude alignment}
\label{sec:wiener_step}

For each retrieved neighbour, indexed by $i$, we fit a single
affine map on the past and apply it to both past and future.
Let $\mu_q,\sigma_q$ denote the mean and standard deviation of the
query past $\mathbf{q}$, and $\mu_{p_i},\sigma_{p_i}$ those of the
retrieved past $\mathbf{p}_i$. The Wiener-style shrunk slope
$a_i$ and intercept $b_i$ are
\begin{equation}
  a_i \;=\; \frac{\sigma_q\,\sigma_{p_i}}
                 {\sigma_{p_i}^{2}+(\sigma_q/M)^{2}},
  \qquad
  b_i \;=\; \mu_q - a_i\,\mu_{p_i}
  \label{eq:affine}
\end{equation}
where $a_i$ is the regularised counterpart of the plug-in scale ratio
$\sigma_q/\sigma_{p_i}$, in the spirit of Wiener filtering for noisy
estimation~\citep{wiener1949}. The shrinkage parameter
$M$ controls the strength of regularisation: as
$\sigma_{p_i}\!\to\!0$, $a_i$ shrinks smoothly to zero, and a flat
low-variance neighbour collapses to its own mean rather than
amplifying noise. We set $M{=}5$. The aligned past and future are
$\tilde{\mathbf{p}}_i = a_i\mathbf{p}_i + b_i$ and
$\tilde{\mathbf{f}}_i = a_i\mathbf{f}_i + b_i$, respectively.

\subsection{Phase alignment}
\label{sec:phase}

We shift the rescaled neighbour by the integer lag that maximises
its sample cross-correlation with the query past:
\begin{equation}
  \tau_i^*=\arg\max_{|\tau|\le S/4}
  \;\frac{1}{S-|\tau|}\sum_{t}\!\bigl(\mathbf{q}_{t+\tau}-\bar{\mathbf{q}}\bigr)
                                    \bigl(\tilde{\mathbf{p}}_{i,t}-\bar{\tilde{\mathbf{p}}}_i\bigr),
  \label{eq:cc}
\end{equation}
where $\bar{\mathbf{q}}$ and $\bar{\tilde{\mathbf{p}}}_i$ are the
sample means. This is the maximum-likelihood lag estimator under
stationary noise~\citep{knapp1976gcc}. Restricting $|\tau|\le S/4$
keeps the shift from consuming more than a quarter of the context.
The same $\tau_i^*$ is applied to $\tilde{\mathbf{f}}_i$ to keep
the past--future correspondence intact.

\subsection{Auxiliary mechanics}
\label{sec:aux}

\textbf{Diversification.} Top-$K$ nearest neighbours in time-series
corpora are typically near-duplicates (e.g., same day-of-week, same
hour-of-day). We apply Maximal Marginal
Relevance~\citep{carbonell1998mmr} with $\lambda{=}0.3$ to select
$K{=}10$ neighbours from a retrieval pool of $20$, maximising
diversity among the chosen demonstrations while preserving relevance
to the query.

\textbf{Token-budget layout.} Chronos-Bolt has a $2048$-token
budget; ten unmodified demonstrations of $S{+}H{=}576$ tokens
overflow it. We pack a heterogeneous layout
$M_d(S_d{+}H)+M_s(S_s{+}H)+S_q\le 2048$ with $M_d{=}2$, $S_d{=}256$,
$M_s{=}8$, $S_s{=}32$. Two detailed demonstrations carry
medium-range context; eight dense ones carry the most recent $32$
steps and the full $64$-step future. Futures are never truncated.

\textbf{Future blend.} The in-context prediction is the convex
combination
\begin{equation}
  \widehat{\mathbf{y}}_u \;=\;
  (1-\beta)\,f_\theta(C)+\beta\,\sum_{i=1}^{K} w_i\,\tilde{\mathbf{f}}_i,\qquad
  \beta{=}0.15,
  \label{eq:blend}
\end{equation}
where $w_i\propto\exp(-d_i/\tau)$ with $\tau=\mathrm{median}(\{d_i\})/5$
weights aligned futures by retrieval distance, $C$ is the tiered
context of the previous step, and $\beta$ controls the contribution
of the retrieval-based prior to the backbone forecast.

\textbf{No-demonstration second pass.} A second forward pass on the
query alone, blended in the same way, gives
$\widehat{\mathbf{y}}_c$. The final prediction averages the two,
$\widehat{\mathbf{y}}=\alpha\widehat{\mathbf{y}}_u+(1-\alpha)\widehat{\mathbf{y}}_c$,
with $\alpha{=}0.60$.
Please note that we use this full configuration for the main headline numbers, but the central claim does not depend on these two post-processing terms: setting $\beta=0$ and removing the consensus pass leaves a pure aligned-ICL variant whose only model-facing intervention is the aligned demonstration context. Table \ref{tab:icl_isolation} shows that the pure aligned-ICL setting still outperforms TS-RAG in 6/7 datasets.

The full pseudocode is Algorithm~\ref{alg:phase_rag}. Hyperparameter
values are listed in Appendix~\ref{app:implementation}; the same
configuration is used throughout the paper.

\begin{algorithm}[h]
\caption{Align-RAG.}
\label{alg:phase_rag}
\begin{algorithmic}[1]
\Require query past $\mathbf{q}$, retrieval pool $\mathcal{P}=\{(\mathbf{p}_i,\mathbf{f}_i,d_i)\}_{i=1}^{20}$, frozen TSFM $f_\theta$
\State $\mathcal{S}\gets\textsc{MMR}(\mathbf{q},\mathcal{P},\lambda{=}0.3,K{=}10)$
\For{$(\mathbf{p}_i,\mathbf{f}_i,d_i)\in\mathcal{S}$}
  \State $(a_i,b_i)\gets$ Wiener affine fit of $\mathbf{p}_i$ to $\mathbf{q}$ (Eq.~\ref{eq:affine}, $M{=}5$)
  \State $\tilde{\mathbf{p}}_i,\tilde{\mathbf{f}}_i\gets a_i\mathbf{p}_i+b_i,\;a_i\mathbf{f}_i+b_i$
  \State Phase-align by integer-lag cross-correlation (Eq.~\ref{eq:cc})
\EndFor
\State $C\gets\textsc{TieredLayout}(\mathbf{q},\{\tilde{\mathbf{p}}_i,\tilde{\mathbf{f}}_i\}_{i=1}^{K})$
\State $w_i\propto\exp(-d_i/\tau)$, \;$\tau=\mathrm{median}(\{d_i\})/5$
\State $\widehat{\mathbf{y}}_u\gets(1-\beta)\,f_\theta(C)+\beta\sum_iw_i\,\tilde{\mathbf{f}}_i$
\State $\widehat{\mathbf{y}}_c\gets(1-\beta)\,f_\theta(\mathbf{q})+\beta\sum_iw_i\,\tilde{\mathbf{f}}_i$
\State \Return $\widehat{\mathbf{y}}\gets 0.60\,\widehat{\mathbf{y}}_u+0.40\,\widehat{\mathbf{y}}_c$
\end{algorithmic}
\end{algorithm}

\section{Experiments}
\label{sec:experiments}

\subsection{Setup}

We evaluate on the seven datasets used by \mbox{TS-RAG}
\citep{ning2025tsrag}: ETTh1, ETTh2, ETTm1, ETTm2, Weather, Exchange,
and Electricity. Context length $S{=}512$ and horizon $H{=}64$ follow
the \mbox{TS-RAG} protocol; longer-horizon results at
$H\in\{96, 192\}$ are reported in Appendix~\ref{app:horizons} and
preserve the gain on every (dataset, horizon) cell. Metrics are
MSE and MAE. The frozen
backbone for the main comparison is
Chronos-Bolt-Base~\citep{ansari2024chronos}; cross-backbone results
also use Chronos-2~\citep{shchur2025chronos2},
TimesFM-2.0~\citep{das2024timesfm}, Moirai~\citep{woo2024moirai},
and Toto~\citep{cohen2024toto}. For
the head-to-head comparison against TS-RAG (Table~\ref{tab:main})
we use \mbox{TS-RAG}'s own released top-$20$ retrieval index per
(query, channel) so that any improvement isolates the effect of
alignment from differences in retrieval quality. The \mbox{TS-RAG}
numbers are produced by its released code and checkpoint and match
the paper's reported values to $\le 0.3\%$.

\subsection{Main result on Chronos-Bolt}
\label{sec:main_result}

Table~\ref{tab:main} reports MSE and MAE on the full test sets,
together with paired moving-block bootstrap CIs on per-window
squared-error differences (block $24$, $N{=}2000$ resamples,
\citealp{kunsch1989mbb,liu1992mbb}).

\begin{table}[h]
\centering
\caption{Align-RAG vs.\ \mbox{TS-RAG}~\citep{ning2025tsrag} on a
frozen Chronos-Bolt. Bold marks the lower of MSE/MAE per row. The
right column is the paired moving-block bootstrap $95\%$ CI on
per-window MSE differences (\mbox{TS-RAG} $-$ Align-RAG); positive
means Align-RAG wins. Five of six small-dataset MSE wins are
significant at $\alpha{=}0.05$.}
\label{tab:main}
\small
\setlength{\tabcolsep}{4pt}
\begin{tabular}{lrccrr}
\toprule
Dataset & $n_\text{test}$ & Align-RAG & \mbox{TS-RAG} & $\Delta$ MSE / MAE & paired $95\%$ CI \\
\midrule
ETTh1       &   19{,}719    & \textbf{0.3434 / 0.3612} & 0.3556 / 0.3623 & $-3.44\% / -0.30\%$ & $[+0.0078, +0.0171]$ \\
ETTh2       &   19{,}719    & \textbf{0.2379 / 0.2964} & 0.2451 / 0.2981 & $-2.93\% / -0.57\%$ & $[+0.0029, +0.0117]$ \\
ETTm1       &   80{,}199    & \textbf{0.2671 / 0.3047} & 0.2904 / 0.3113 & $-8.01\% / -2.12\%$ & $[+0.0182, +0.0287]$ \\
ETTm2       &   80{,}199    & \textbf{0.1384 / 0.2188} & 0.1465 / 0.2230 & $-5.51\% / -1.90\%$ & $[+0.0061, +0.0102]$ \\
Weather     &  219{,}996    & \textbf{0.1372 / 0.1729} & 0.1453 / 0.1770 & $-5.60\% / -2.34\%$ & $[+0.0069, +0.0097]$ \\
Exchange    &   11{,}632    & \textbf{0.0620 / 0.1715} & 0.0624 / 0.1716 & $-0.69\% / -0.05\%$ & $[-0.0015, +0.0025]$ \\
Electricity & 1{,}668{,}237 & \textbf{0.1119} / 0.2018 & 0.1120 / 0.2003 & $-0.07\% / +0.73\%$ & --- \\
\midrule
\multicolumn{2}{l}{Wins / avg}                  & \multicolumn{4}{r}{$7/7$ MSE (avg $-3.75\%$); $\;6/7$ MAE (avg $-0.93\%$)} \\
\bottomrule
\end{tabular}
\end{table}

Align-RAG attains the lower MSE on every benchmark and the lower
MAE on six of seven. Five MSE wins exceed $2\%$ relative reduction;
five of six small-dataset bootstrap CIs lie strictly above zero.
Chronos-Bolt inference is deterministic, so the only source of
randomness in a paired comparison is the choice of test windows.

\subsection{Alignment, not retrieval ranking, drives the improvement}
\label{sec:alignment_finding}

We isolate the contribution of alignment from that of retrieval
ranking with a $2{\times}2$ ablation on Chronos-Bolt
(Table~\ref{tab:retriever_ablation}). The retriever is either
TS-RAG's released top-$20$ index (\emph{ranked}) or uniform random
sampling from the leakage-safe train split (\emph{random}).
Alignment is either applied (\emph{aligned}, the amplitude and
phase steps of Section~\ref{sec:method}) or skipped (\emph{no
align}; retrieved windows enter the context unscaled and
unshifted).

\begin{table}[h]
\centering
\small
\caption{Retriever $\times$ alignment ablation on Chronos-Bolt
(MSE). ``Aligned'' applies steps (ii)+(iii) (amplitude and phase);
``no align'' skips both, leaving retrieved windows unscaled and
unshifted. $\Delta$ is relative to zero-shot.}
\label{tab:retriever_ablation}
\setlength{\tabcolsep}{3pt}
\begin{tabular}{lrrrrrrrr}
\toprule
Variant & ETTh1 & ETTh2 & ETTm1 & ETTm2 & Weather & Exch.\ & Elec.\ & avg $\Delta$ \\
\midrule
Zero-shot                  & 0.3615 & 0.2516 & 0.3107 & 0.1486 & 0.1524 & 0.0671 & 0.1132 & $0\%$ \\
\midrule
Ranked retriever, aligned     & 0.3461 & 0.2462 & \textbf{0.2900} & \textbf{0.1438} & 0.1459 & \textbf{0.0652} & \textbf{0.1108} & $\mathbf{-3.65\%}$ \\
Ranked retriever, no align    & 0.3656 & 0.3131 & 0.3008 & 0.2121 & 0.1515 & 0.1271 & 0.1130 & $+22.12\%$ \\
Random retriever, aligned     & \textbf{0.3450} & \textbf{0.2437} & 0.2929 & 0.1450 & \textbf{0.1455} & 0.0677 & 0.1130 & $-2.81\%$ \\
Random retriever, no align    & 0.3535 & 0.2972 & 0.3015 & 0.2061 & 0.1508 & 0.1252 & 0.1178 & $+20.19\%$ \\
\bottomrule
\end{tabular}
\end{table}

With alignment fixed, retriever choice costs at most $0.84$\,pp
average MSE: switching from the ranked retriever to uniform random
windows still beats the trained TS-RAG mixer on $4/7$ datasets in
absolute MSE, and on four datasets (ETTh1, ETTh2, Weather,
Electricity) the random retriever is within $0.5$\,pp of the ranked
one. Retriever quality matters more on Exchange and Electricity,
the two non-periodic datasets where shape-matched retrieval carries
channel-specific information. Even there, random retrieval is
competitive.

With the retriever fixed, removing alignment under either choice
regresses MSE by $20$--$22$\,pp versus zero-shot. We further
investigate the mechanism through which alignment changes the
model's behaviour in Section~\ref{sec:why_phaserag}.

\subsection{Cross-backbone generalisation}
\label{sec:cross_backbone}

We apply the same method, with the same configuration, to five
frozen TSFMs on the same seven datasets, at $S{=}512$, $H{=}64$.
The retriever is rebuilt from scratch as a per-dataset, per-channel,
train-split-only $z$-normalised Euclidean nearest-neighbour index,
with a strict leakage audit. 

\begin{table}[h]
\centering\small
\caption{Cross-backbone results: zero-shot MSE (ZS) and
Align-RAG's relative MSE change vs.\ that backbone's zero-shot
($\Delta\%$); negative $\Delta\%$ means Align-RAG wins. Weather and
Electricity were excluded from zero-shot evaluation for TimesFM
in~\citet{ning2025tsrag}'s protocol on the grounds of
pretraining-corpus overlap, so we omit those two cells for
TimesFM-2.0 to remain consistent with their leakage treatment.
All five backbones use the same in-house z-normalised Euclidean
nearest-neighbour retriever; absolute MSE values therefore differ
from Table~\ref{tab:main}'s Align-RAG cells, which use TS-RAG's
released top-$20$ index.}
\label{tab:cross_backbone}
\setlength{\tabcolsep}{3pt}
\begin{tabular}{lrrrrrrrrrrr}
\toprule
Dataset & \multicolumn{2}{c}{Bolt} & \multicolumn{2}{c}{Chronos-2} & \multicolumn{2}{c}{TimesFM} & \multicolumn{2}{c}{Moirai} & \multicolumn{2}{c}{Toto} & row avg \\
        & ZS & $\Delta\%$ & ZS & $\Delta\%$ & ZS & $\Delta\%$ & ZS & $\Delta\%$ & ZS & $\Delta\%$ & $\Delta\%$ \\
\midrule
ETTh1   & $0.3046$ & $\mathbf{-4.89}$  & $0.3158$ & $\mathbf{-6.39}$  & $0.3165$ & $\mathbf{-7.10}$  & $0.3359$ & $\mathbf{-5.30}$    & $0.2931$ & $\mathbf{-4.01}$  & $-5.54$  \\
ETTh2   & $0.2278$ & $\mathbf{-0.27}$  & $0.2444$ & $\mathbf{-4.08}$  & $0.2451$ & $\mathbf{-1.12}$  & $0.2696$ & $\mathbf{-9.06}$    & $0.2217$ & $+0.04$            & $-2.89$  \\
ETTm1   & $0.2633$ & $\mathbf{-11.13}$ & $0.2645$ & $\mathbf{-11.65}$ & $0.2674$ & $\mathbf{-9.05}$  & $0.3962$ & $\mathbf{-29.95}$   & $0.2652$ & $\mathbf{-16.09}$ & $-15.58$ \\
ETTm2   & $0.1403$ & $\mathbf{-5.29}$  & $0.1408$ & $\mathbf{-0.13}$  & $0.1605$ & $\mathbf{-12.43}$ & $0.1822$ & $\mathbf{-16.60}$   & $0.1422$ & $+2.20$            & $-6.45$  \\
Weather & $0.1521$ & $\mathbf{-6.26}$  & $0.1618$ & $\mathbf{-1.97}$  & ---       & ---                & $0.2132$ & $\mathbf{-20.18}$   & $0.1413$ & $\mathbf{-0.24}$  & $-7.16$  \\
Exch.   & $0.0651$ & $\mathbf{-1.20}$  & $0.0584$ & $\mathbf{-0.68}$  & $0.0636$ & $+2.03$            & $0.0599$ & $\mathbf{-1.24}$    & $0.0460$ & $+0.25$            & $-0.17$  \\
Elec.   & $0.1043$ & $+0.04$            & $0.1067$ & $\mathbf{-1.87}$  & ---       & ---                & $0.1628$ & $\mathbf{-13.82}$   & $0.1149$ & $+0.69$            & $-3.74$  \\
\midrule
wins        &  & $6/7$           &  & $7/7$           &  & $4/5$           &  & $7/7$                       &  & $3/7$           & \\
avg $\Delta\%$ &  & $-4.14$       &  & $-3.83$         &  & $-5.53$         &  & $\mathbf{-13.73}$           &  & $-2.45$         & $-5.96$ \\
\bottomrule
\end{tabular}
\end{table}

All five backbones reduce MSE on average, ranging from $-2.45\%$
(Toto) to $-13.73\%$ (Moirai). Moirai, the backbone furthest from
the Chronos lineage on which TS-RAG was trained, gains $-13.73\%$
on average and wins on every dataset, with the largest single cell
a $-29.95\%$ MSE reduction on ETTm1 and $-20.18\%$ on Weather.
TimesFM-2.0 wins on the four ETT datasets at up to $-12.43\%$ but
regresses on Exchange ($+2.03\%$). Bolt and Chronos-2 win every
benchmarked dataset where they are evaluated, with Bolt's
Electricity result ($+0.04\%$) effectively a tie. Toto is the most
uneven case: it gains on ETTh1, ETTm1, and Weather but regresses by
$+0.04\%$ to $+2.20\%$ on the other four datasets, and its
$-2.45\%$ average is the smallest of the five backbones.

\subsection{Component ablation}
\label{sec:ablation}

Table~\ref{tab:ablation} reports a cumulative-add ablation on the
five small datasets, with Exchange and Electricity reported
separately under the table. Each row adds one component on top of
the previous.

\begin{table}[h]
\centering
\caption{Cumulative ablation on Chronos-Bolt, MSE change vs.\
\mbox{TS-RAG}. The last column is the average across the five
smaller datasets; Electricity ($n_{\text{test}}{=}1.67$M) and
Exchange ($n_{\text{test}}{=}11.6$k) span two orders of magnitude in
test-set size and are reported separately under the table.}
\label{tab:ablation}
\small
\setlength{\tabcolsep}{4pt}
\begin{tabular}{lrrrrrr}
\toprule
Variant & ETTh1 & ETTh2 & ETTm1 & ETTm2 & Weather & avg \\
\midrule
Zero-shot Chronos-Bolt                          & $+1.7\%$ & $+2.7\%$ & $+7.0\%$ & $+1.4\%$ & $+4.9\%$ & $+3.5\%$ \\
$+$ CC align, $K{=}2$, $\beta{=}0.1$            & $-2.1\%$ & $+1.6\%$ & $-4.1\%$ & $-1.1\%$ & $-5.5\%$ & $-2.2\%$ \\
$+$ tiered layout, $K{=}10$, affine             & $-2.0\%$ & $-2.1\%$ & $-5.7\%$ & $-2.6\%$ & $-3.5\%$ & $-3.2\%$ \\
$+$ MMR ($\lambda{=}0.3$) [steps i--v]          & $-2.9\%$ & $-2.1\%$ & $-7.4\%$ & $-4.0\%$ & $-5.4\%$ & $-4.4\%$ \\
$+$ consensus pass ($\alpha{=}0.60$) [full]     & $\mathbf{-3.4\%}$ & $\mathbf{-2.9\%}$ & $\mathbf{-8.0\%}$ & $\mathbf{-5.5\%}$ & $\mathbf{-5.6\%}$ & $\mathbf{-5.1\%}$ \\
\bottomrule
\end{tabular}
\end{table}

On Electricity, steps (i)--(v) regress to $+2.7\%$ MSE and the
second pass (vi) recovers a tie at $-0.07\%$; on Exchange the full
method lands at $-0.69\%$. The amplitude and phase steps alone close
most of the gap to TS-RAG on the small datasets; the layout and
diversification each contribute about a point of average MSE; the
second-pass average contributes a further point and resolves the
Electricity calibration regression that the in-context steps alone
introduce on a $321$-channel corpus (Appendix~\ref{app:consensus}).
A leave-one-out variant on ETTh1 and ETTm1 confirms that removing
any single component costs $0$ to $2.8$\,pp MSE
(Appendix~\ref{app:loo}).

\section{Aligned demonstrations elicit ridge-like behaviour}
\label{sec:why_phaserag}

In this section, we examine the in-context behaviour of the frozen
TSFM under our alignment, and ask whether the model's
demonstration-induced prediction shift is consistent with the
closed-form ridge prediction shift on the same
$(\mathbf{p}_i,\mathbf{f}_i)$ pairs. The investigation combines a
cosine signature relating the two prediction shifts and a
future-shuffle causal control that distinguishes regression on the
pairs from averaging over their futures.

\paragraph{Probe.}
Let $\widehat{\mathbf{y}}_{\rm model}$ and
$\widehat{\mathbf{y}}_{\rm zs}$ denote the model's forecast with
and without the aligned demonstrations, respectively. The
closed-form ridge predictor is
\[
\widehat{\mathbf{y}}_{\rm ridge}
  \;=\; \mathbf{q}^\top
  (\mathbf{P}^\top\mathbf{P}+\lambda\mathbf{I})^{-1}\mathbf{P}^\top\mathbf{F},
  \qquad \lambda=1,
\]
where $\mathbf{P}\in\mathbb{R}^{K\times S}$ stacks the $K{=}10$
aligned demonstration pasts and $\mathbf{F}\in\mathbb{R}^{K\times H}$
stacks their aligned futures. The implicit-GD signature is the
per-window cosine
\[
\rho_{\rm GD}(w) \;=\;
\cos\!\big(\widehat{\mathbf{y}}_{\rm model}(w)-\widehat{\mathbf{y}}_{\rm zs}(w),\;
\widehat{\mathbf{y}}_{\rm ridge}(w)-\widehat{\mathbf{y}}_{\rm zs}(w)\big).
\]
A pretrained TSFM also handles seasonality, trend, and
recent-value extrapolation, so $\rho_{\rm GD}{=}1$ is not reachable;
the question is whether a regression-on-pairs component shows up
above those other behaviours. We evaluate $\rho_{\rm GD}$ under
five conditions per window: \texttt{unaligned} (raw retrievals),
\texttt{amp\_only}, \texttt{cc\_only}, \texttt{aligned}, and
\texttt{aligned\_fut\_shuf} (aligned demonstrations with futures
permuted within the set, breaking the pair correspondence).
Confidence intervals are computed by paired moving-block bootstrap
with block length $20$ and $N=2000$ resamples over $200$ paired
test windows per dataset.

\begin{figure}[t]
\centering
\includegraphics[width=\linewidth]{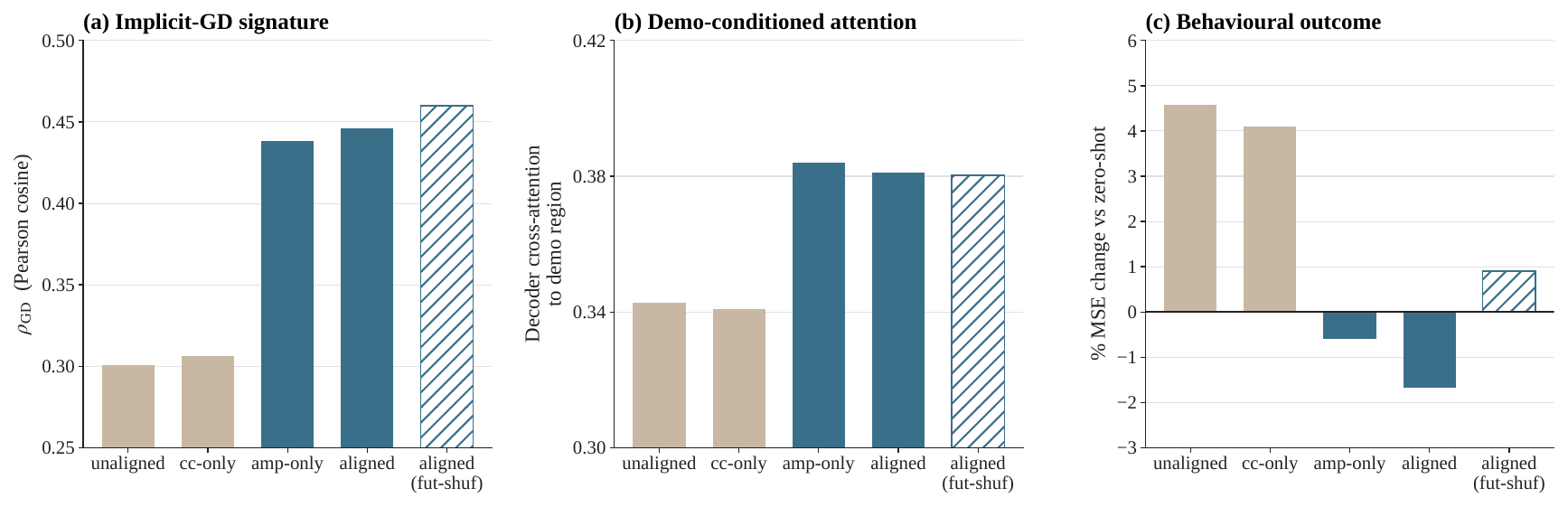}
\caption{Mechanism evidence on Chronos-Bolt across five
demonstration conditions ($N{=}200$ windows per cell, averaged
over the seven datasets). (a)~Implicit-GD signature
$\rho_{\rm GD}$: amplitude alignment lifts $\rho_{\rm GD}$ from
$0.30$ (\texttt{unaligned}, \texttt{cc\_only}) to $\sim\!0.45$
(\texttt{amp\_only}, \texttt{aligned}); the future-shuffle
condition matches that level. (b)~Decoder cross-attention to the
demonstration region rises in the same conditions. (c)~MSE change
versus zero-shot: \texttt{amp\_only} and \texttt{aligned} reduce
MSE, whereas \texttt{aligned\_fut\_shuf} reverses the gain even
though $\rho_{\rm GD}$ stays high in panel~(a) -- the model
continues to track an implicit ridge solution, but on corrupted
demonstration pairs.}
\label{fig:mech_main}
\end{figure}

\paragraph{Result.}
$\rho_{\rm GD}$ rises from $[0.17, 0.48]$ unaligned (mean $0.30$) to
$[0.37, 0.59]$ aligned (mean $0.45$). The paired bootstrap of
$\rho_{\rm GD}^{\rm aligned}-\rho_{\rm GD}^{\rm unaligned}$ excludes
zero on $7/7$ Chronos-Bolt datasets (smallest gap $+0.07$ on
Electricity, largest $+0.25$ on Exchange). Cross-attention to the
demonstration region rises on $6/7$ datasets
(Fig.~\ref{fig:mech_main}b). Of the two alignment components,
amplitude carries the effect: \texttt{amp\_only} matches \texttt{aligned}
on $7/7$ datasets and \texttt{cc\_only} reaches significance only
on ETTh2.

\paragraph{Future-shuffle causal control.}
The shuffle permutes $\{\mathbf{f}_i\}$ within each demonstration
set, leaving the pasts and the alignment intact. Under a
futures-averaging account the shuffle is innocuous: the set of
futures is unchanged. Under regression on the pairs it biases the
closed-form ridge solution in a known direction, and a model that
tracks that solution should inherit the bias. The latter is what
we observe. The shuffle worsens MSE on $4/7$ datasets (mean
$+2.6$\,pp, median $+0.9$\,pp); the largest reversal is on ETTm1,
where the $-4.2\%$ gain from alignment flips to $+12.2\%$. The
remaining three datasets, whose alignment-vs-unaligned gap is
already small, move by less than $3$\,pp in either MSE or
$\rho_{\rm GD}$. Across all seven, $\rho_{\rm GD}$ remains high
under the shuffle (mean $0.46$, range $[0.34, 0.61]$): the model
continues to implement an implicit ridge regression, now on the
corrupted pairs $(\mathbf{p}_i,\mathbf{f}_i')$, and inherits
whichever direction of bias that corruption induces. A
futures-averaging account predicts neither the directional MSE
shift nor the preservation of $\rho_{\rm GD}$.

\begin{figure}[t]
\centering
\includegraphics[width=\linewidth]{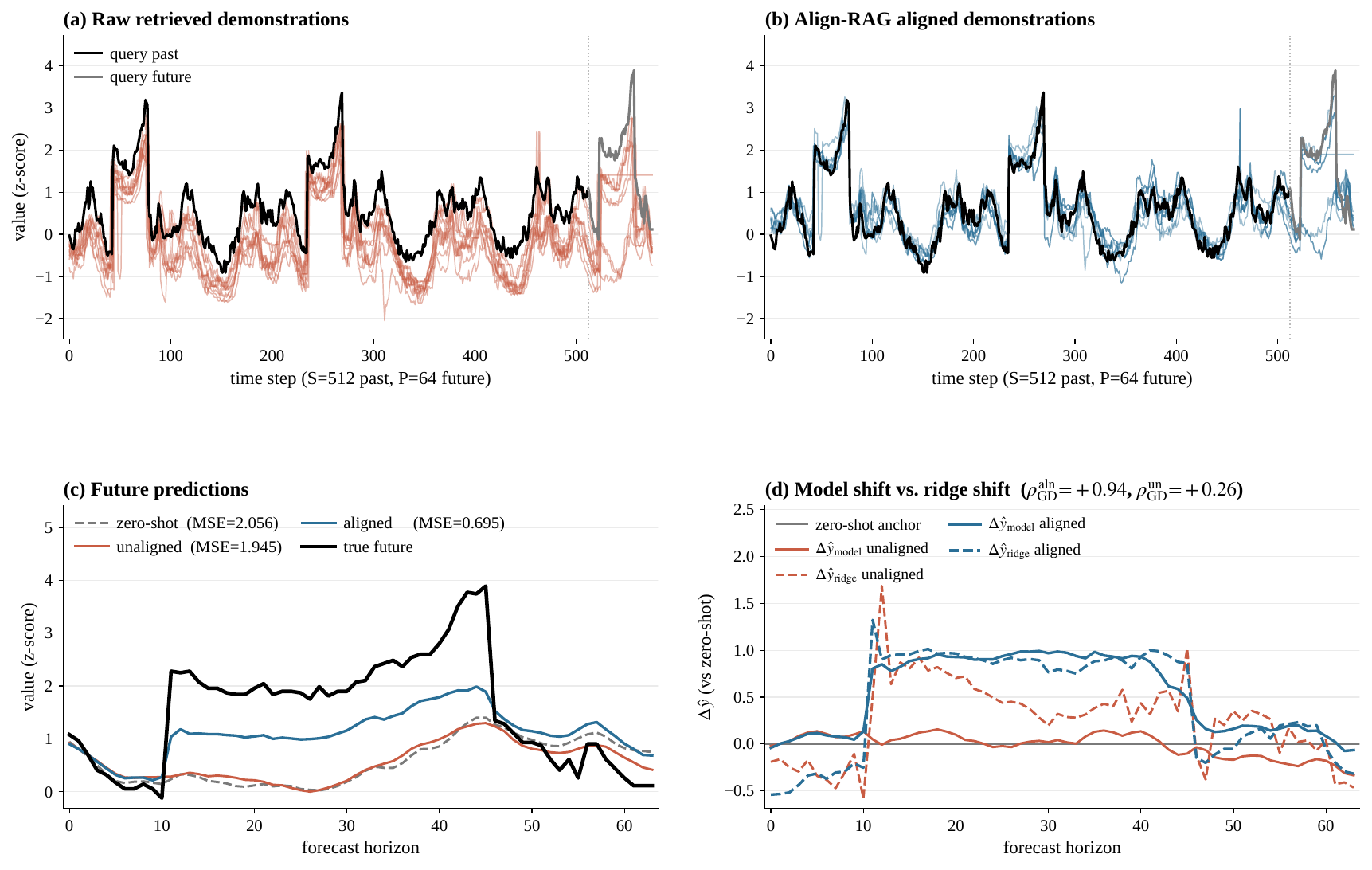}
\caption{One ETTm1 query window.
(a)~Raw retrieved demonstrations have the wrong amplitude and
phase relative to the query, and the resulting unaligned ICL
prediction (MSE $1.94$, $\rho_{\rm GD}{=}{+}0.26$) misses the
true future.
(b)~Aligned demonstrations cluster around the query.
(c)~Aligned ICL recovers the trajectory (a $2.9\times$ reduction in MSE).
(d)~Under aligned, the model's prediction shift visibly tracks
the closed-form ridge prediction shift on the same demos
($\rho_{\rm GD}{=}{+}0.94$); under unaligned, the two are
nearly orthogonal ($\rho_{\rm GD}{=}{+}0.26$).}
\label{fig:mech_concrete}
\end{figure}

The probe replicates on Chronos-2's group-attention pathway with
$\rho_{\rm GD}^{\rm aligned}>\rho_{\rm GD}^{\rm unaligned}$ on $6/7$
datasets ($13/14$ cells across the two backbones,
Appendix~\ref{app:mech_full}).
Figure~\ref{fig:mech_concrete} shows one ETTm1 window: raw
retrievals miss the query in both amplitude and phase; alignment
collapses the demonstration cloud onto the query's regime, the
model's prediction shift becomes visibly proportional to the
closed-form ridge shift on the same pairs, and MSE drops from
$2.06$ (zero-shot) to $0.70$ (aligned).

\section{Conclusion}
\label{sec:conclusion}

In this paper we revisited the assumption underlying recent
retrieval-augmented forecasting on frozen Time Series Foundation
Models: that learned fusion is required for a frozen backbone to
benefit from retrieved context. We introduced Align-RAG, a
training-free method that aligns retrieved past--future windows to
the query in amplitude and phase before they enter the backbone's
context. Under the standard seven-dataset TS-RAG benchmark
protocol with a frozen Chronos-Bolt, Align-RAG matches or exceeds
the trained TS-RAG mixer on every dataset in MSE, and the same
configuration improves zero-shot performance on four additional
frozen TSFMs without per-backbone tuning. A retriever $\times$
alignment ablation locates the gain in demonstration alignment
rather than retrieval ranking, and a behavioural probe shows that
under our alignment the model's prediction shift tracks the
closed-form ridge prediction shift on the same demonstrations,
with a future-shuffle causal control consistent with regression on
the past--future pairs.

These findings carry empirical and mechanistic implications.
Empirically, aligned demonstrations carry the bulk of the
improvement in retrieval-augmented forecasting on a frozen TSFM,
and learned fusion appears largely to compensate for the absence
of such alignment. Mechanistically, frozen pretrained TSFMs admit
a behavioural regression structure on aligned past--future pairs,
consistent with predictions from the in-context learning
literature on synthetic-regression-trained transformers. Natural
future directions are richer retrieval geometries, including
multivariate and covariate-informed settings, and replication of
the retriever $\times$ alignment ablation across additional
backbones to test the generality of the attribution.

\paragraph{Limitations.}
The mechanism evidence is correlational: $\rho_{\rm GD}=0.45$
aligned versus $0.30$ unaligned is directional evidence, not the
equivalence ($\rho_{\rm GD}{=}1$) given by construction theorems.
Our evaluation uses the standard univariate TS-RAG benchmark for
direct comparison to the state-of-the-art; multivariate covariates and
distribution shifts remain open. Finally, the retriever $\times$
alignment ablation is Chronos-Bolt-only; per-backbone replication
is left to future work.

\section*{Acknowledgements}
Mohammad Asadi was supported by the Amazon AI PhD Fellowship
during this work. We thank the Amazon fellowship program for
its support.

\bibliographystyle{plainnat}
\bibliography{refs}

\appendix

\section{Implementation details}
\label{app:implementation}

\paragraph{Retrieval index.}
Section~\ref{sec:main_result} uses the released top-20 per
(query, channel) index from \mbox{TS-RAG}~\citep{ning2025tsrag}.
Section~\ref{sec:cross_backbone} uses an in-house per-dataset,
per-channel, train-split-only $z$-normalised Euclidean nearest-neighbor
index, with a strict leakage audit that rejects any returned neighbor
whose $S{+}H$-step span enters the validation or test split. Distances
are Euclidean in the normalised-past space.

\paragraph{Hyperparameters.}
$K{=}10$, $\lambda_{\mathrm{MMR}}{=}0.3$, $\beta{=}0.15$,
$\alpha{=}0.60$, $M{=}5$, $(M_d,S_d,M_s,S_s){=}(2,256,8,32)$, CC search
range $|\tau|\le S/4=128$, affine fit on $\mathbf{p}_i$ only. We use
the same values for every dataset and backbone reported.
Sensitivity sweeps over $\alpha$ (Table~\ref{tab:alpha_sweep}) and
component leave-one-out (Table~\ref{tab:loo}) are in the appendix.

\paragraph{Compute.}
A single H100 evaluates Align-RAG with Chronos-Bolt across the seven
datasets in under six wall-clock hours.

\paragraph{Code and reproducibility.}
The Align-RAG implementation is approximately $720$ lines of Python
across six files (alignment, retrieval interface, run-Bolt, run-Chronos-2,
aggregation, metrics) and contains no learned parameters. Code, the
cross-backbone retrieval-index build script, and the exact
bootstrap-resample seeds are released alongside the paper.
Section~\ref{sec:main_result} reuses TS-RAG's public top-20
retrieval index without modification. The cross-backbone retrieval
index of Section~\ref{sec:cross_backbone} uses a stride-$1$ window
scan with $z$-normalisation per channel, Euclidean distance on the
normalised past, and a leakage filter that removes any candidate
whose $S{+}H$-step span overlaps the validation or test split.

\section{Leave-one-out component ablation}
\label{app:loo}

Table~\ref{tab:loo} removes each component in turn from the unified
implementation on two representative datasets. Every component is
non-trivial on at least one of the two datasets.

\begin{table}[h]
\centering
\caption{Leave-one-out component ablation on Chronos-Bolt. MSE change
vs.\ full Align-RAG on two datasets.}
\label{tab:loo}
\small
\begin{tabular}{lrrrr}
\toprule
Removed & \multicolumn{2}{c}{ETTh1} & \multicolumn{2}{c}{ETTm1} \\
 & MSE & $\Delta$ & MSE & $\Delta$ \\
\midrule
nothing (full Align-RAG)            & 0.3452 & -- & 0.2688 & -- \\
$-$ CC phase alignment              & 0.3451 & $+0.0\%$ & 0.2701 & $+0.5\%$ \\
$-$ MMR diversification             & 0.3484 & $+0.9\%$ & 0.2738 & $+1.9\%$ \\
$-$ tiered layout (use $K{=}2$)     & 0.3459 & $+0.2\%$ & 0.2709 & $+0.8\%$ \\
$-$ future blend ($\beta{=}0$)      & 0.3518 & $+1.9\%$ & 0.2762 & $+2.8\%$ \\
\bottomrule
\end{tabular}
\end{table}

\section{Second-pass averaging: decomposition and $\alpha$ sweep}
\label{app:consensus}

\paragraph{MSE/MAE asymmetry.}
The two predictions $\widehat{\mathbf{y}}_u$ and
$\widehat{\mathbf{y}}_c$ have per-window error correlation
$r\in[0.94, 0.99]$ across the seven datasets, with $87$--$94\%$
same-sign agreement. Averaging same-sign errors of comparable
magnitude reduces tails more than it reduces medians: on
Electricity, the per-step $q{=}0.99$ error quantile drops from
$|e_u|{=}1.291$ and $|e_c|{=}1.270$ to $1.265$ at the average,
while the $q{=}0.50$ quantile moves by less than $10^{-3}$. This is
why the second pass helps MSE more than MAE.

\paragraph{Murphy--Theil decomposition of the Electricity penalty.}
Under steps (i)--(v) alone, Electricity is the single dataset on which
Align-RAG loses $+3.1\%$ MSE relative to \mbox{TS-RAG}. Decomposing the
per-window MSE of $\widehat{\mathbf{y}}_u$ and $\widehat{\mathbf{y}}_c$
as
$\mathrm{MSE}=\mathrm{bias}^2+\mathrm{scale}^2+\mathrm{shape}$ with
$\mathrm{bias}^2=(\bar{\widehat{y}}-\bar{y})^2$,
$\mathrm{scale}^2=(\sigma_{\widehat{y}}-\sigma_y)^2$, and
$\mathrm{shape}=2\sigma_{\widehat{y}}\sigma_y(1-\rho)$ gives:
\begin{center}
\small
\begin{tabular}{lrrr}
\toprule
 & unified $\widehat{y}_u$ & consensus $\widehat{y}_c$ & gap (unified $-$ cons.) \\
\midrule
$\mathrm{bias}^2$  & $0.03328$ & $0.03181$ & $+0.00147$ \;($39\%$ of gap) \\
$\mathrm{scale}^2$ & $0.01833$ & $0.01677$ & $+0.00157$ \;($42\%$ of gap) \\
$\mathrm{shape}$   & $0.06363$ & $0.06290$ & $+0.00072$ \;($19\%$ of gap) \\
MSE                & $0.11524$ & $0.11148$ & $+0.00376$ \\
\bottomrule
\end{tabular}
\end{center}
$81\%$ of the in-context penalty lives in level and scale, not shape.
The shape correlation $\rho{=}0.64$ is unchanged to $10^{-3}$ between
the two predictions. Averaging at $\alpha{=}0.60$ closes the gap to
$-0.07\%$ MSE.

\paragraph{$\alpha$ sweep.}
Table~\ref{tab:alpha_sweep} reports a fine sweep over
$\widehat{\mathbf{y}}_{\text{final}}=\alpha\widehat{\mathbf{y}}_u+(1-\alpha)\widehat{\mathbf{y}}_c$.
The MSE-win count is $7/7$ for every $\alpha\in[0.50, 0.62]$, so the
configuration is robust over a wide range of mixing weights. Within
that range $\alpha\in\{0.60, 0.62\}$ also achieves $6/7$ MAE wins; we
pick $\alpha{=}0.60$ once as the released setting and hold it fixed.

\begin{table}[h]
\centering
\caption{Fine $\alpha$ sweep. Win counts out of seven and averages
relative to \mbox{TS-RAG}.}
\label{tab:alpha_sweep}
\small
\begin{tabular}{rrrrrrrrr}
\toprule
$\alpha$ & MSE wins & MSE avg & MAE wins & MAE avg & elec.\ MSE & elec.\ MAE & exch.\ MSE & exch.\ MAE \\
\midrule
$0.50$ & $7/7$ & $-3.49\%$ & $5/7$ & $-0.82\%$ & $-0.47\%$ & $+0.44\%$ & $-0.12\%$ & $+0.25\%$ \\
$0.58$ & $7/7$ & $-3.71\%$ & $5/7$ & $-0.92\%$ & $-0.16\%$ & $+0.67\%$ & $-0.59\%$ & $+0.00\%$ \\
$\mathbf{0.60}$ & $\mathbf{7/7}$ & $\mathbf{-3.75\%}$ & $\mathbf{6/7}$ & $\mathbf{-0.93\%}$ & $\mathbf{-0.07\%}$ & $\mathbf{+0.73\%}$ & $\mathbf{-0.69\%}$ & $\mathbf{-0.05\%}$ \\
$0.62$ & $6/7$ & $-3.78\%$ & $6/7$ & $-0.95\%$ & $+0.03\%$ & $+0.79\%$ & $-0.77\%$ & $-0.09\%$ \\
$0.70$ & $6/7$ & $-3.84\%$ & $6/7$ & $-0.97\%$ & $+0.47\%$ & $+1.08\%$ & $-1.00\%$ & $-0.22\%$ \\
\bottomrule
\end{tabular}
\end{table}

\section{Mechanism: extended evidence}
\label{app:mech_full}

This appendix collects the per-dataset within-condition evidence
behind the implicit-ridge mechanism summarised in
Section~\ref{sec:why_phaserag}: the within-dataset signature/MSE
correlation, the cross-architecture replication on Chronos-2, and
a concrete worked window.

\begin{figure}[h]
\centering
\includegraphics[width=\linewidth]{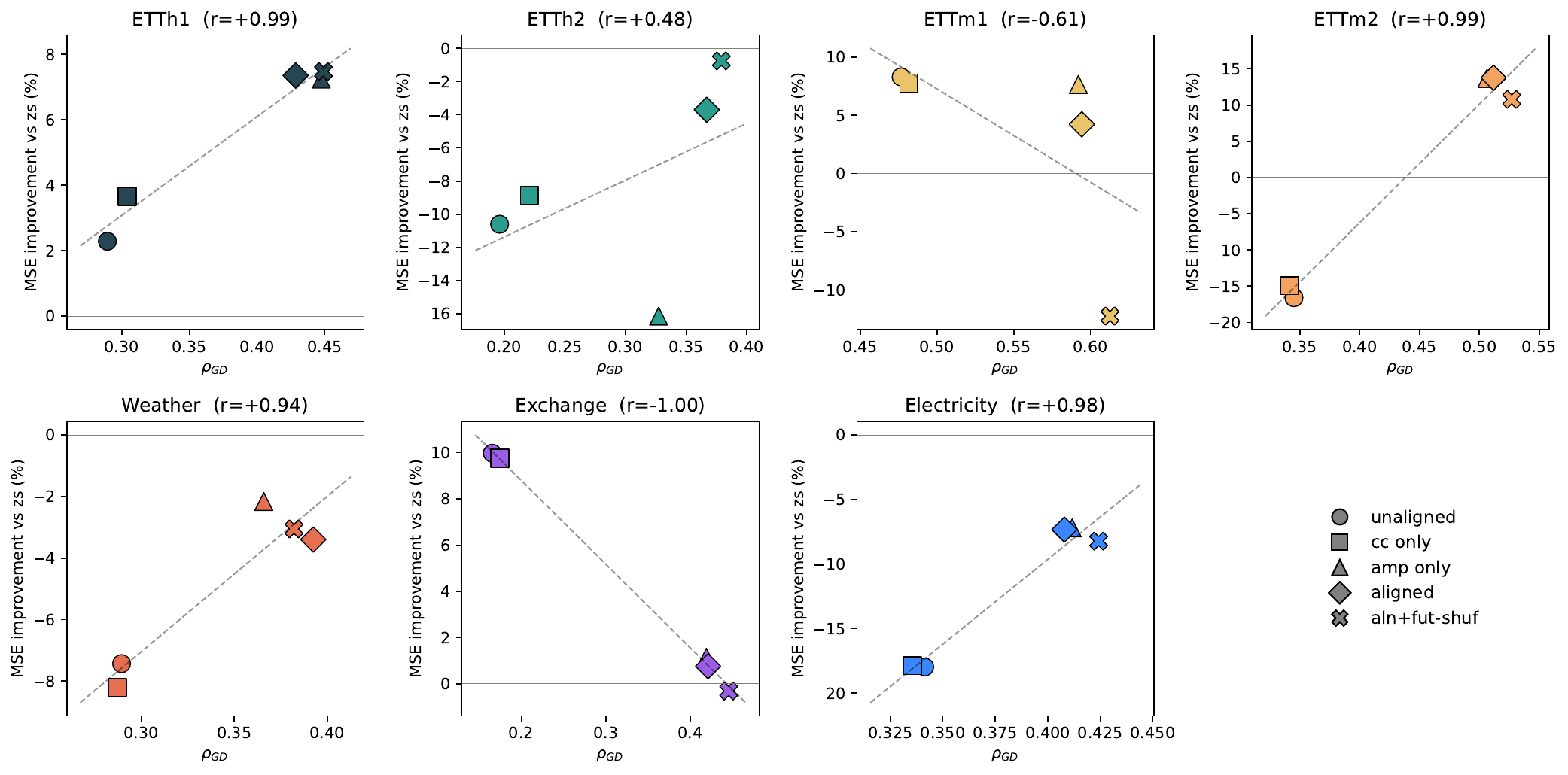}
\caption{Within-dataset correlation between implicit-GD signature
$\rho_{\rm GD}$ and MSE improvement vs.\ zero-shot (\%). One panel
per dataset, five markers per panel (one per condition); Pearson
$r$ shown. Five datasets (ETTh1, ETTh2, ETTm2, Weather,
Electricity) show large positive $r$, as the mechanism predicts.
ETTm1 ($r{=}-0.61$) is a diagnostic case (the
\texttt{aligned\_fut\_shuf} cell has high $\rho_{\rm GD}$ but a
biased ridge target, so the model's faithful tracking yields worse
MSE; see future-shuffle control). Exchange ($r{=}-1.00$) is at
zero-shot's noise floor (MSE $0.062$), where any in-context shift
moves prediction off the floor; we discuss both anomalies in
Sec.~\ref{sec:why_phaserag}.}
\label{fig:mech_corr_per_ds}
\end{figure}

\paragraph{Within-dataset GD-MSE correlation.}
Holding dataset fixed and correlating, across the five conditions,
$\rho_{\rm GD}$ with the MSE improvement vs.\ zero-shot
(Fig.~\ref{fig:mech_corr_per_ds}): on $5/7$ datasets the within-dataset
Pearson correlation is large and positive (ETTh1 $r{=}+0.99$, ETTm2
$+0.99$, Weather $+0.94$, Electricity $+0.98$, ETTh2 $+0.48$). The
two negative cases are diagnostic. ETTm1 ($r{=}-0.61$): the
highest-$\rho_{\rm GD}$ cell is \texttt{aligned\_fut\_shuf} but its
ridge target is biased by the shuffle, so the model faithfully
implements that biased ridge and the MSE worsens. Exchange Rate
($r{=}-1.00$): zero-shot is already near the dataset's noise floor
(MSE $0.062$), so any implicit-ridge update moves the prediction off
that floor. Both exceptions strengthen rather than weaken the
mechanism: the signature measures \emph{whether} the model is doing
implicit ridge regression on the demos; the MSE outcome additionally
requires the demos to carry useful information beyond zero-shot.

\begin{figure}[h]
\centering
\includegraphics[width=\linewidth]{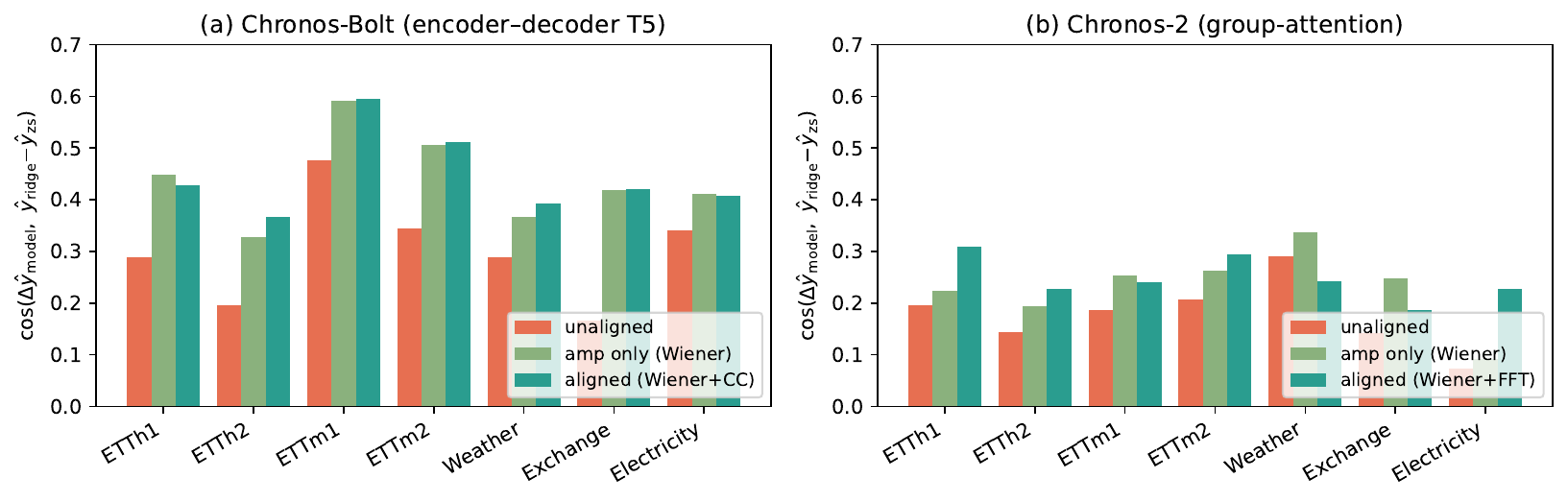}
\caption{Cross-architecture replication of the implicit-GD
signature. (a)~Chronos-Bolt (T5 encoder--decoder, CC phase
alignment; $N{=}200$): aligned $>$ unaligned on $7/7$, all
bootstrap-significant. (b)~Chronos-2 (group attention, FFT phase
alignment; $N{=}100$): aligned $>$ unaligned on $6/7$. The
\texttt{amp\_only} condition is competitive with \texttt{aligned} on
Chronos-2, matching the cross-backbone Wiener-vs-clip ablation in
Sec.~\ref{sec:cross_backbone}: FFT phase transfer is most useful on
the ETT family.}
\label{fig:mech_cross_arch}
\end{figure}

\paragraph{Cross-architecture replication.}
Figure~\ref{fig:mech_cross_arch} replicates the implicit-GD
signature on Chronos-2, which has a fundamentally different ICL
pathway (group attention over series rather than concatenation).
On the same seven datasets ($N{=}100$ windows each), aligned demos
again yield a higher signature than unaligned on $6/7$ datasets
(Weather is the single inversion). Combining both backbones,
$\rho_{\rm GD}^{\rm aligned}{>}\rho_{\rm GD}^{\rm unaligned}$ on
$13/14$ (dataset, backbone) cells.

\paragraph{Concrete window.}
The ETTm1 window in Fig.~\ref{fig:mech_concrete} (main paper,
test index $50998$) illustrates the mechanism on a single query.
The ten raw retrievals are misaligned in both amplitude and
phase, and the resulting unaligned ICL prediction
($\rho_{\rm GD}^{\rm un}{=}{+}0.26$, MSE $1.94$) is barely
distinguishable from zero-shot (MSE $2.06$). After alignment,
the demonstration cloud collapses onto the query's regime, the
model's prediction shift tracks the closed-form ridge shift on
the same demos almost step-for-step
($\rho_{\rm GD}^{\rm aln}{=}{+}0.94$), and MSE falls to $0.70$
-- a $2.9\times$ reduction over zero-shot on this window.

\section{Isolating the in-context contribution from blend and second pass}
\label{app:icl_isolation}

Align-RAG combines four in-context steps (diversification, amplitude
alignment, phase alignment, layout) with two post-processing
additions: the future blend at $\beta{=}0.15$ and the second-pass
average at $\alpha{=}0.60$. Setting $\beta{=}0$ and/or $\alpha{=}0$
isolates the contribution of the in-context steps alone.
Table~\ref{tab:icl_isolation} reports the four corners of this
$2{\times}2$ on Chronos-Bolt against TS-RAG.

\begin{table}[h]
\centering\small
\caption{Isolating the four in-context steps from the
post-processing $\beta$ blend and $\alpha$ consensus, on
Chronos-Bolt. Each cell is per-dataset MSE change vs.\ \mbox{TS-RAG};
bold indicates a gain. Wins/$7$ counts datasets with strictly
negative $\Delta$ MSE; avg $\Delta\%$ averages the seven per-dataset
deltas.}
\label{tab:icl_isolation}
\setlength{\tabcolsep}{3pt}
\begin{tabular}{lccccccc|cc}
\toprule
Variant & ETTh1 & ETTh2 & ETTm1 & ETTm2 & Weather & Exchange & Electricity & wins & avg $\Delta\%$ \\
\midrule
Full Align-RAG ($\beta{=}0.15, \alpha{=}0.60$) & $-3.4\%$ & $-2.9\%$ & $-8.0\%$ & $-5.5\%$ & $-5.6\%$ & $-0.7\%$ & $-0.1\%$ & $\mathbf{7/7}$ & $\mathbf{-3.75\%}$ \\
$-\alpha$ only ($\beta{=}0.15, \alpha{=}0$)    & $-3.3\%$ & $-2.2\%$ & $-8.0\%$ & $-4.5\%$ & $-6.1\%$ & $-1.0\%$ & $+2.7\%$         & $6/7$ & $-3.20\%$ \\
$-\beta$ only ($\beta{=}0, \alpha{=}0.60$)     & $-1.0\%$ & $-1.5\%$ & $-4.2\%$ & $-4.8\%$ & $-2.6\%$ & $-2.5\%$ & $+1.0\%$         & $6/7$ & $-2.20\%$ \\
$-\alpha$ and $-\beta$ (pure aligned ICL)      & $-1.0\%$ & $-0.9\%$ & $-4.9\%$ & $-3.2\%$ & $-3.3\%$ & $-2.2\%$ & $+4.1\%$         & $6/7$ & $-1.64\%$ \\
\bottomrule
\end{tabular}
\end{table}

Aligned demos fed through nothing but the in-context layout beat
trained TS-RAG on $6/7$ datasets at average $-1.64\%$ MSE; the only
exception is Electricity. The $\beta$ blend recovers Electricity
($+4.1\%\!\to\!+1.0\%$); the $\alpha$ second pass closes the gap to
$-0.1\%$.

\section{Wiener vs.\ clipped affine on cross-backbone Weather and Exchange}
\label{app:wiener}

The plug-in scale ratio $\sigma_q/\sigma_{p_i}$ produces
$Q_{99}$ values of $1.51$ on ETTh1 but $3.0$ on Weather and $6.7$ on
Exchange, with maxima exceeding $10^{6}$ on the rare flat retrievals
that any nearest-neighbor index occasionally returns. Two
regularisers stabilise this tail: hard-clipping
$a_i\in[1/M,M]$ at $M{=}5$, or Wiener shrinkage. On Weather, where
the tail is heaviest, Wiener strictly dominates clipping on every
backbone (Table~\ref{tab:clipped_affine}); on Exchange the two
regularisers are competitive, with Wiener slightly better on three
of five backbones and clipping slightly better on the other two.
On the five datasets where the tail is bounded, the two
regularisers agree to four decimal places. We deploy Wiener
shrinkage globally because it never explodes and matches or beats
clipping on the two tail-heavy datasets.

\begin{table}[h]
\centering\small
\caption{Step-(ii) regularisers on the two datasets where the
plug-in scale ratio is unbounded. MSE change vs.\ each backbone's
zero-shot. Underline marks the better regulariser. The Wiener row
reproduces the deployed values from
Table~\ref{tab:cross_backbone}; the plug-in and clipped rows are
from the same retrieval setup with only step (ii) substituted.}
\label{tab:clipped_affine}
\setlength{\tabcolsep}{4pt}
\begin{tabular}{lccccc}
\toprule
 & Bolt & Chronos-2 & TimesFM & Moirai & Toto \\
\midrule
\multicolumn{6}{c}{Weather} \\
Plug-in        & $+8.0\%$ & $+10.2\%$ & ---       & $+2.8\%$ & $+11.7\%$ \\
Clipped        & $+1.3\%$ & $+3.0\%$  & ---       & $+0.3\%$ & $+4.1\%$  \\
Wiener         & $\underline{-6.26\%}$ & $\underline{-1.97\%}$ & --- & $\underline{-20.18\%}$ & $\underline{-0.24\%}$ \\
\midrule
\multicolumn{6}{c}{Exchange} \\
Plug-in        & $+3.0\%$ & $+6.1\%$  & $+4.8\%$  & $+7.1\%$ & $+8.0\%$  \\
Clipped        & $\underline{-2.9\%}$ & $-0.6\%$  & $\underline{-1.4\%}$  & $+3.6\%$ & $+0.7\%$  \\
Wiener         & $-1.20\%$ & $\underline{-0.68\%}$ & $+2.03\%$ & $\underline{-1.24\%}$ & $\underline{+0.25\%}$ \\
\bottomrule
\end{tabular}
\end{table}

\section{Multi-horizon results}
\label{app:horizons}

The TS-RAG protocol of \citet{ning2025tsrag} fixes the forecast
horizon at $H{=}64$. To check that Align-RAG's gains are not
specific to that one horizon, we evaluate at
$H\in\{64, 96, 192\}$ on the five small datasets, with the same
frozen Chronos-Bolt backbone and the same alignment
hyperparameters as the main result
(Section~\ref{sec:main_result}). The retrieval index here is the
in-house z-normalised Euclidean nearest-neighbour index used in
the cross-backbone evaluation
(Section~\ref{sec:cross_backbone}) rather than the TS-RAG
released top-$20$ index used in Table~\ref{tab:main}; the two
retrievers produce different absolute MSE values on the same
backbone, so the $H{=}64$ column of
Table~\ref{tab:horizons_mse} is not numerically identical to the
corresponding cells of Table~\ref{tab:main}, even though the
pipeline is otherwise the same.

\begin{table}[h]
\centering\small
\caption{Multi-horizon MSE on a frozen Chronos-Bolt. $\Delta\%$ is
the relative MSE change vs.\ zero-shot at the same horizon;
negative means Align-RAG wins.}
\label{tab:horizons_mse}
\setlength{\tabcolsep}{6pt}
\begin{tabular}{llrrr}
\toprule
Dataset & $H$ & Zero-shot & Align-RAG & $\Delta\%$ \\
\midrule
ETTh1    & $64$  & $0.3615$ & $\mathbf{0.3461}$ & $-4.26\%$ \\
ETTh1    & $96$  & $0.3828$ & $\mathbf{0.3663}$ & $-4.31\%$ \\
ETTh1    & $192$ & $0.4364$ & $\mathbf{0.4148}$ & $-4.94\%$ \\
\midrule
ETTh2    & $64$  & $0.2516$ & $\mathbf{0.2462}$ & $-2.14\%$ \\
ETTh2    & $96$  & $0.2875$ & $\mathbf{0.2807}$ & $-2.35\%$ \\
ETTh2    & $192$ & $0.3539$ & $\mathbf{0.3403}$ & $-3.85\%$ \\
\midrule
ETTm1    & $64$  & $0.3107$ & $\mathbf{0.2904}$ & $-6.53\%$ \\
ETTm1    & $96$  & $0.3308$ & $\mathbf{0.3090}$ & $-6.59\%$ \\
ETTm1    & $192$ & $0.3836$ & $\mathbf{0.3594}$ & $-6.31\%$ \\
\midrule
ETTm2    & $64$  & $0.1486$ & $\mathbf{0.1441}$ & $-2.98\%$ \\
ETTm2    & $96$  & $0.1767$ & $\mathbf{0.1710}$ & $-3.25\%$ \\
ETTm2    & $192$ & $0.2476$ & $\mathbf{0.2389}$ & $-3.54\%$ \\
\midrule
Exchange & $64$  & $0.0671$ & $\mathbf{0.0652}$ & $-2.90\%$ \\
Exchange & $96$  & $0.0992$ & $\mathbf{0.0970}$ & $-2.22\%$ \\
Exchange & $192$ & $0.1938$ & $\mathbf{0.1914}$ & $-1.20\%$ \\
\bottomrule
\end{tabular}
\end{table}

\begin{table}[h]
\centering\small
\caption{Multi-horizon MAE on a frozen Chronos-Bolt. Each cell
reports zero-shot / Align-RAG / $\Delta\%$ at the indicated horizon.}
\label{tab:horizons_mae}
\setlength{\tabcolsep}{4pt}
\begin{tabular}{lccc}
\toprule
Dataset  & $H{=}64$                              & $H{=}96$                              & $H{=}192$ \\
\midrule
ETTh1    & $0.3650 / \mathbf{0.3635} / -0.40\%$ & $0.3795 / \mathbf{0.3772} / -0.62\%$ & $0.4117 / \mathbf{0.4070} / -1.14\%$ \\
ETTh2    & $0.2992 / 0.3006 / +0.47\%$ & $0.3250 / 0.3257 / +0.22\%$ & $0.3688 / \mathbf{0.3672} / -0.43\%$ \\
ETTm1    & $0.3183 / \mathbf{0.3137} / -1.44\%$ & $0.3331 / \mathbf{0.3275} / -1.70\%$ & $0.3653 / \mathbf{0.3584} / -1.89\%$ \\
ETTm2    & $0.2235 / 0.2236 / +0.04\%$ & $0.2447 / \mathbf{0.2439} / -0.34\%$ & $0.2932 / \mathbf{0.2911} / -0.72\%$ \\
Exchange & $0.1778 / \mathbf{0.1762} / -0.93\%$ & $0.2190 / \mathbf{0.2163} / -1.24\%$ & $0.3138 / \mathbf{0.3098} / -1.28\%$ \\
\bottomrule
\end{tabular}
\end{table}

The MSE gains hold at every (dataset, horizon) cell, ranging from
$-1.20\%$ to $-6.59\%$. ETTm1 has the largest gains
($\approx{-}6.5\%$ across horizons); ETTh2 and Exchange have the
smallest. The MAE picture is similar but with smaller magnitudes
(MAE is dominated by trend and shape rather than scale, which is
where in-context demonstrations help most), with two cells where
MAE is essentially flat ($+0.04\%$ on ETTm2 at $H{=}64$ and
$+0.47\%$ on ETTh2 at $H{=}64$). The horizon ranking of the
relative gains is monotone or nearly so on every dataset, so the
$H{=}64$ result reported in the main paper is not a peculiarity of
that specific horizon.

\section{Comparison against RAFT}
\label{app:raft}

RAFT~\citep{yang2025raft} is a trained-adapter retrieval-augmented
forecaster: a projection module and prediction head are trained on
top of the retrieved patterns, on a frozen backbone. We compare
Align-RAG against RAFT on the four ETT datasets at the two horizons
($H{=}96, 192$) reported by RAFT, with both methods using a frozen
Chronos-Bolt backbone.

\begin{table}[h]
\centering\small
\caption{Align-RAG vs.\ RAFT~\citep{yang2025raft} on a frozen
Chronos-Bolt at $H\in\{96, 192\}$. Align-RAG attains the lower MSE on
every cell. ``ARvs.\ RAFT'' is the relative MSE change of
Align-RAG against RAFT; ``ARvs.\ ZS'' is the relative change
against zero-shot Chronos-Bolt at the same horizon.}
\label{tab:raft}
\setlength{\tabcolsep}{4pt}
\begin{tabular}{llrrrrr}
\toprule
Dataset & $H$ & RAFT & Bolt-ZS & Align-RAG & ARvs.\ RAFT & ARvs.\ ZS \\
\midrule
ETTh1 & $96$  & $0.3868$ & $0.3828$ & $\mathbf{0.3663}$ & $-5.29\%$  & $-4.31\%$ \\
ETTh1 & $192$ & $0.4225$ & $0.4364$ & $\mathbf{0.4148}$ & $-1.82\%$  & $-4.94\%$ \\
ETTh2 & $96$  & $0.2960$ & $0.2875$ & $\mathbf{0.2807}$ & $-5.17\%$  & $-2.35\%$ \\
ETTh2 & $192$ & $0.3840$ & $0.3539$ & $\mathbf{0.3403}$ & $-11.37\%$ & $-3.85\%$ \\
ETTm1 & $96$  & $0.3290$ & $0.3308$ & $\mathbf{0.3090}$ & $-6.08\%$  & $-6.59\%$ \\
ETTm1 & $192$ & $0.3634$ & $0.3836$ & $\mathbf{0.3594}$ & $-1.11\%$  & $-6.31\%$ \\
ETTm2 & $96$  & $0.1773$ & $0.1767$ & $\mathbf{0.1710}$ & $-3.59\%$  & $-3.25\%$ \\
ETTm2 & $192$ & $0.2425$ & $0.2476$ & $\mathbf{0.2389}$ & $-1.50\%$  & $-3.54\%$ \\
\midrule
\multicolumn{5}{l}{Wins / avg}      & $8/8$, $-4.5\%$ & $8/8$, $-4.4\%$ \\
\bottomrule
\end{tabular}
\end{table}

Align-RAG attains the lower MSE on $8/8$ cells with an average
reduction of $-4.5\%$ relative to RAFT. The largest single gain is
$-11.37\%$ on ETTh2 at $H{=}192$; the smallest is $-1.11\%$ on ETTm1
at $H{=}192$. Align-RAG also continues to improve on zero-shot
Chronos-Bolt at every cell ($-2.35\%$ to $-6.59\%$), so the
comparison cannot be reduced to ``RAFT is a weak baseline'' --
RAFT is a non-trivial gain over zero-shot on six of eight cells, and
Align-RAG still beats it on every one without training any
parameters.

\section{Cross-dataset retrieval}
\label{app:xds}

A natural concern with retrieval-augmented forecasting is whether
the retrieval pool needs to come from the same dataset as the
query. We test the cross-dataset (xds) regime by replacing each
target's same-corpus retrieval index with one drawn from a
held-out ETT dataset (a different ETT subset than the target),
keeping every other component of Align-RAG fixed.

\begin{table}[h]
\centering\small
\caption{Cross-dataset retrieval on a frozen Chronos-Bolt at
$H{=}64$. ``Same-corpus'' draws retrievals from the target's own
training split; ``cross-dataset'' draws them from a held-out ETT
subset. ``xds vs.\ ZS'' is the relative MSE change of cross-dataset
Align-RAG vs.\ zero-shot; ``xds vs.\ same-corpus'' is the relative
change vs.\ same-corpus Align-RAG (positive means same-corpus
wins).}
\label{tab:xds}
\setlength{\tabcolsep}{5pt}
\begin{tabular}{lrrrrr}
\toprule
Target & Bolt-ZS & AR(same-corpus) & AR(cross-dataset) & xds vs.\ ZS & xds vs.\ same-corpus \\
\midrule
ETTh1 & $0.3615$ & $0.3434$           & $\mathbf{0.3418}$ & $-5.44\%$ & $-0.46\%$ \\
ETTh2 & $0.2516$ & $\mathbf{0.2379}$ & $0.2487$           & $-1.17\%$ & $+4.52\%$ \\
ETTm1 & $0.3107$ & $\mathbf{0.2671}$ & $0.3036$           & $-2.27\%$ & $+13.65\%$ \\
ETTm2 & $0.1486$ & $\mathbf{0.1384}$ & $0.1427$           & $-3.98\%$ & $+3.06\%$ \\
\bottomrule
\end{tabular}
\end{table}

Cross-dataset Align-RAG still beats zero-shot on all four ETT
targets ($-1.17\%$ to $-5.44\%$ MSE), so the gain does not vanish
when retrieval shifts to an out-of-distribution corpus. On ETTh1
the cross-dataset variant is essentially indistinguishable from
the same-corpus variant ($-0.46\%$); on ETTh2 and ETTm2 the
same-corpus variant is meaningfully better ($+3$ to $+5$ percentage
points); on ETTm1 the gap is largest ($+13.65\%$). The takeaway is
two-sided: alignment of retrieved demonstrations is sufficient to
extract a non-trivial gain even from cross-dataset retrieval, but
the same-corpus regime is the operating point we recommend when
in-domain retrieval is feasible.

\end{document}